\documentclass[runningheads]{llncs}

\usepackage[mobile]{eccv}

\usepackage{xcolor}         % colors
\usepackage{tabularx}
\usepackage{subcaption}
\usepackage{multicol}
\usepackage{multirow}
\usepackage{tabularray}
\usepackage{amsmath}
\usepackage{amssymb}
\usepackage{colortbl}
\usepackage{placeins}

\usepackage{eccvabbrv}

\usepackage{graphicx}
\usepackage{booktabs}

\usepackage[accsupp]{axessibility}  % Improves PDF readability for those with disabilities.

\usepackage{hyperref}

\usepackage{orcidlink}

\definecolor{cvprblue}{rgb}{0.21,0.49,0.74}
\definecolor{pltblue}{RGB}{174, 199, 232}
\definecolor{pltorange}{RGB}{255, 229, 204}
\definecolor{pltgreen}{RGB}{204, 229, 204}
\definecolor{pltred}{RGB}{229, 204, 204}
\definecolor{pltpurple}{RGB}{239, 218, 230}

\definecolor{tabblue}{HTML}{1f77b4}
\definecolor{taborange}{HTML}{ff7f0e}
\definecolor{tabgreen}{HTML}{2ca02c}
\definecolor{tabred}{HTML}{d62728}
\definecolor{tabpurple}{HTML}{9467bd}
\definecolor{cblue}{RGB}{173, 201, 233}
\definecolor{clblue}{RGB}{222, 234, 246}
\definecolor{corange}{RGB}{255, 152, 67}
\definecolor{lorgange}{RGB}{255, 221, 149}
\definecolor{myred}{RGB}{174,66,38}

\newcolumntype{C}{>{\centering\arraybackslash}X}

\newcommand{\cc}[1]{\cellcolor{clblue!50}{#1}}

\begin{document}

\title{Leveraging Multimodal Large Language Models for All-in-One Image Restoration via a Mixture of Frequency Experts} 

\titlerunning{MMFE-IR}

\author{Eunho Lee\inst{1} \and
Rei Kawakami\inst{2} \and
Youngbae Hwang\inst{1}}

\authorrunning{Eunho Lee et al.}

\institute{Chungbuk National University, Cheongju 28644, Republic of Korea\\
\email{\{ehlee, ybhwang\}@cbnu.ac.kr}\\
\and
Institute of Science Tokyo, Tokyo 152-8550, Japan\\
\email{reikawa@sc.eng.isct.ac.jp}}

\maketitle

\begin{abstract}
All-in-one image restoration seeks to recover clean images from inputs affected by diverse and unknown degradations using a unified framework.
%All-in-one image restoration aims to recover clean images degraded by diverse and unknown factors within a unified framework. 
Recent methods have shown strong performance by identifying degradation characteristics to guide the restoration process.
%and restoring images using a single network. 
%Recent approaches have shown remarkable performance by identifying degradation characteristics and restoring images within a single network. 
However, many of them treat degradations as discrete categories, which limits their ability to model the continuous relational structure that arises in composite degradations.
%However, they often treat degradations as discrete categories, limiting their ability to model the complex and distributed dependencies present in composite degradations.
To address this issue, we propose a multimodal large language model (MLLM)-guided image restoration framework that exploits multimodal embeddings as guidance for low-level restoration. Specifically, MLLM-derived features are injected into an encoder-decoder architecture through an MLLM-guided fusion block (MGFB) to enhance degradation-aware representations. In addition, we incorporate a mixture-of-frequency-experts (MoFE) module that adaptively combines frequency experts using MLLM-guided contextual cues. To further improve expert routing, we design an MLLM-guided router with a relational alignment loss that encourages routing patterns consistent with the embedding-space relationships of degraded inputs.
Extensive experiments on multiple benchmarks show that the proposed method achieves strong performance across diverse restoration settings and establishes a new state of the art on the challenging CDD11 dataset, outperforming previous methods by up to 1.35 dB.
%
%In contrast, multimodal large language models (MLLMs) encode rich cross-modal knowledge from large-scale pretraining, which can better capture these semantic relationships among degradations.
%To leverage these capabilities, we propose an MLLM-guided image restoration framework that integrates these high-level knowledge into low-level processing. MLLM-derived features are fused into an encoder-decoder architecture via an MLLM-guided fusion block (MGFB) to learn degradation-aware representations, while a mixture-of-frequency-experts (MoFE) module adaptively combines experts based on contextual guidance from the MLLM.
%An MLLM-guided router, optimized with an MLLM-guided loss, dynamically assigns expert weights according to the semantic dependencies among degradations. 
%Extensive experiments demonstrate that our method achieves state-of-the-art performance across multiple benchmarks, outperforming prior approaches by up to 1.35 dB on the challenging CDD11 dataset, which contains both single and composite degradations.
  \keywords{Image Restoration \and Mixture of Experts \and Multimodal Large Language Model}
\end{abstract}

% ==================== Introduction ====================
\section{Introduction}
Image restoration is a fundamental low-level vision task that aims to reconstruct a clean, high-quality image from a given input corrupted by diverse real-world degradations.
Conventional methods are typically designed to handle a single type of degradation, such as noise~\cite{guo2019toward, zamir2020cycleisp}, blur~\cite{zhang2020deblurring, gao2024efficient}, haze~\cite{ren2018gated, ren2020single}, or rain~\cite{ren2019progressive, jie2023imagederaining}. 
Each degradation requires a separately trained model or dedicated parameters, which limits scalability and generalization. 

\begin{figure}
    \centering
    \includegraphics[width=\linewidth]{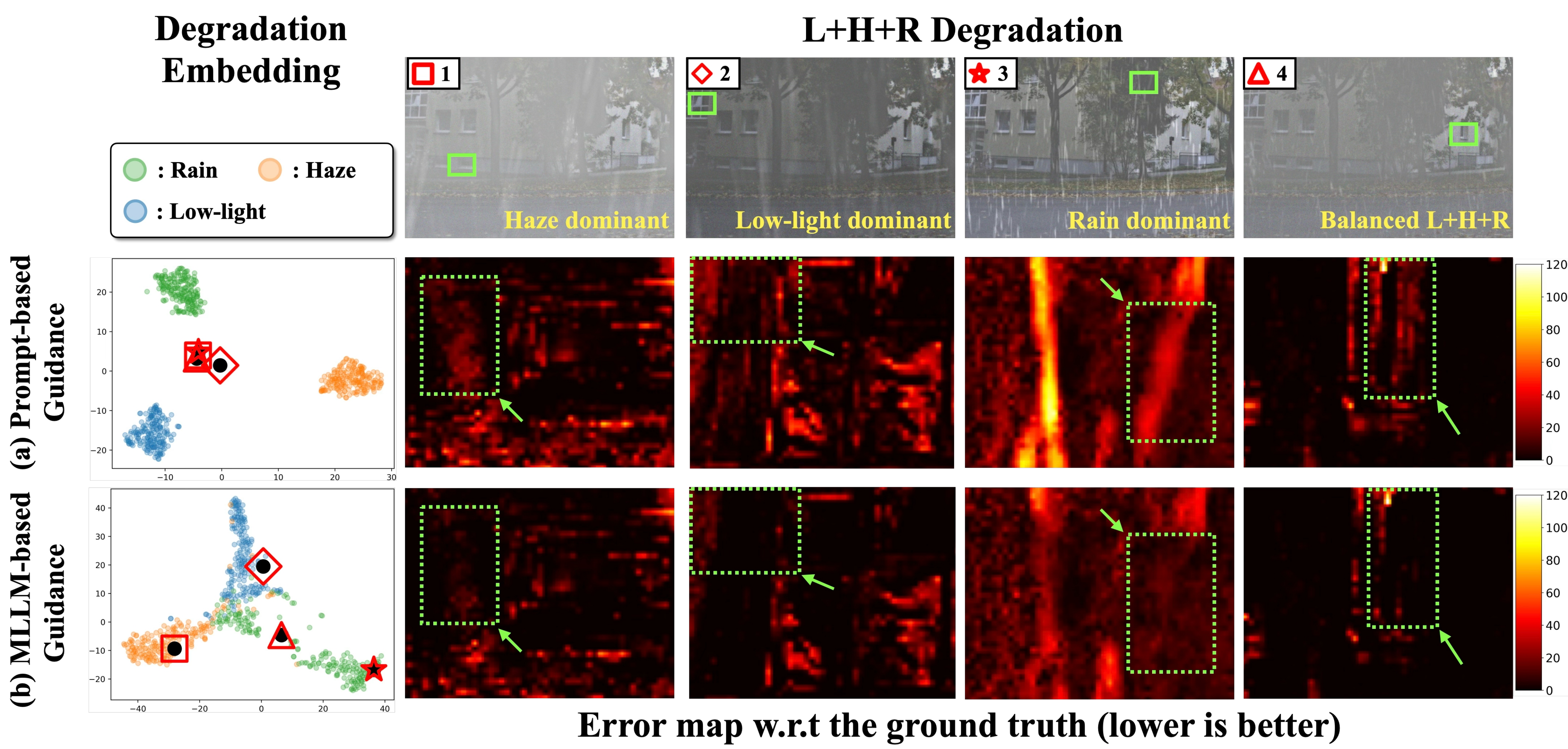}
    \caption{
 \textbf{Comparison of latent embeddings for composite degradation (Low-light + Haze + Rain).}
The prompt-based embedding often collapse mixed degradations into a single cluster, disregarding variations in dominant characteristics.
In contrast, the pretrained MLLM embedding distribute samples across a semantic space, positioning composite cases according to their dominant degradation (e.g., haze-, low-light-, or rain-dominant), thereby more accurately capturing the structure of mixed degradations.
}
\vspace{-0.5cm}
    \label{fig:latent}
\end{figure}

To address these limitations, all-in-one image restoration methods~\cite{li2022airnet, valanarasu2022transweather, zhang2023idr} attempt to handle various degradation types within a unified framework.
These methods typically embed degradation-aware priors into the network to guide the restoration process, often treating degradations as discrete types to be identified or disentangled.
%primarily focusing on separating distinct degradation.
Such priors are derived from degradation-sensitive representations, which are obtained either through contrastive learning to learn discriminative features or through prompt-based approaches that leverage textual information~\cite{li2022airnet, potlapalli2023promptir, conde2024instructir}.

However, these approaches do not fully capture the coexistence and interaction among degradations.
\cref{fig:latent} illustrates diverse composite-degradation scenarios, including the case of haze-, rain-, low-light-dominant and balanced mixtures.
Degradation-sensitive representations in existing methods~\cite{guo2024onerestore} tend to partition degradation types into isolated, discrete clusters, as they are often learned via contrastive objectives or explicit classification.
This separation overlooks the continuous nature of degradation characteristics: even under composite degradations, if one degradation is dominant, its characteristics become more salient. (e.g., haze-dominant L+H+R cases are closer to haze than balanced mixtures).
As a result, such representation-based guidance cannot adequately capture changes in degradation composition, often leading to suboptimal restoration for composite cases (\cref{fig:latent}a).

Recently, multimodal large language models (MLLMs) \cite{liu2024llava_vl, chen2024internvl, peng2024qwen2_VL, bai2025qwen2_5} have shown impressive performance across a wide range of vision-language tasks. These models are pre-trained on large-scale web image-text datasets such as LAION~\cite{schuhmann2022laion}, which include diverse real-world degradations. They are then refined through visual instruction tuning~\cite{liu2024llava_vl} on complex multimodal data for tasks such as visual question answering (VQA), captioning, and reasoning.
This training pipeline encourages MLLMs to learn rich cross-modal associations, enabling their embedding space to represent degradations as a continuous spectrum, as illustrated in \cref{fig:latent}. Such a property directly addresses the challenge of discrete categorization.

Nonetheless, leveraging MLLMs' knowledge for low-level vision remains challenging.
Existing attempts in image restoration often rely solely on high-level textual representations~\cite{he2024training_free, chen2024restoreagent, ai2024dreamclear, zhu2025an}, largely underutilizing the rich internal visual knowledge of MLLMs learned from large-scale web image-text data.
This results in a representation gap, as coarse textual representations provide insufficient fine-grained cues to guide accurate pixel-level reconstruction.

To address this, our framework leverages both high-level textual features and MLLMs' internal visual knowledge.
First, to incorporate MLLM-derived visual cues into a low-level restoration network, we introduce an MLLM-guided fusion block (MGFB).
MGFB fuses MLLM-derived vision embedding with vision-language-conditioned embeddings within the network's encoder and decoder blocks, enhancing the encoder's feature extraction and the decoder's detail-preserving reconstruction.
This explicit guidance enables the restoration network to better exploit the visual knowledge derived from MLLM during restoration.

While leveraging MLLMs' high-level textual semantics to better handle diverse degradations, we propose a mixture-of-frequency-experts (MoFE) module and further introduce an MLLM-guided router optimization strategy.
Since different degradations exhibit distinct frequency characteristics~\cite{cui2025adair}, we design MoFE to operate in the frequency domain rather than the spatial domain.
% MoFE consists of frequency-specialized experts, each capturing complementary frequency components; by dynamically composing these experts, the model can consider diverse degradations.
MoFE consists of frequency-specialized experts, each capturing complementary frequency components; the router leverages the MLLM textual representation to predict their mixture weights for each input, enabling dynamic composition to handle diverse degradations.
To enable expert composition, we propose an MLLM-guided loss that supervises the MoFE router using the MLLM's continuous textual priors.
Overall, by jointly leveraging MLLMs' internal visual knowledge and high-level textual semantics, our framework mitigates the representation gap and learns distributed degradation relationships, leading to effective restoration under complex degradations (\cref{fig:latent}b).

Our main contributions are summarized as follows:
\begin{itemize}
\item
% An MLLM-guided fusion block (MGFB) is proposed to effectively bridge the representation gap by integrating both vision-derived and vision-language-conditioned embeddings into the restoration network.
%We leverage pretrained MLLMs’ continuous degradation embeddings to overcome the limitations imposed by the rigid discretization of prompt-based representations, allowing more accurate capture of the complex structure of degradations.
We leverage pretrained MLLMs’ \emph{continuous degradation embeddings} to overcome limitations of rigid prompt discretization, enabling more accurate capture of the complex structure of degradations.
\item
% A mixture-of-frequency-experts (MoFE) module is proposed to handle diverse degradations by employing frequency-specialized experts.
An MLLM-guided fusion block (MGFB) is introduced to bridge the representation gap by integrating MLLM vision embeddings and vision-language-conditioned embeddings into the restoration network.
\item 
% We introduce an MLLM-guided loss that trains the MoFE router to learn the scattered relationships among degradations, thereby mitigating the limitation of prior methods that treat them as discrete categories.
We propose a mixture-of-frequency-experts (MoFE) module with an MLLM-guided router, trained via an MLLM-guided loss to learn distributed degradation relationships, thereby improving restoration under diverse and composite degradations.
\end{itemize}

The proposed method achieves state-of-the-art results on various all-in-one image restoration benchmarks and shows strong generalization to composite degradations, outperforming previous approaches by up to 1.35 dB on the challenging CDD11 dataset~\cite{guo2024cdd11}.

% ==================== Related works ====================
\section{Related Work}
\noindent
\textbf{All-in-one Image Restoration.}
Image restoration aims to recover clean images from inputs corrupted by various degradations. 
Early approaches focused on task-specific networks designed for individual degradation types.
Some incorporated learning-based priors tailored for specific degradations like denoising, dehazing, or deblurring~\cite{zhang2017learning, guo2019toward, zhang2020deblurring, zhang2018densely}.
Others focused on network architectures designed to capture task-specific characteristics~\cite{ren2018gated, ren2019progressive, ren2020single}.
More recently, single powerful network architectures, such as convolutional neural networks (CNNs)~\cite{zhang2018residual, liu2019dual, zamir2021mprnet, chen2022nafnet} and Transformer based-models~\cite{liang2021swinir, wang2022uformer, zamir2022restormer}, have achieved dominant performance when trained for these individual tasks separately.
However, such task-specific designs require training separate models for different degradation types, which limits scalability in real-world scenarios where multiple degradations occur.

Recent studies have explored all-in-one image restoration frameworks that handle diverse degradation types within a unified model~\cite{li2022airnet, zhang2023idr, valanarasu2022transweather, park2023all}.
AirNet~\cite{li2022airnet} addresses multiple degradations by leveraging contrastive learning to capture degradation-aware representations. Following this, IDR~\cite{zhang2023idr} integrates task-specific priors into a unified representation using dynamic routing and learnable PCA.
MoE-based approaches~\cite{zamfir2025moceir, zamfir2024daair} have further been proposed for all-in-one image restoration, introducing adaptive expert architectures that dynamically allocate computational capacity or model specialization according to degradation complexity within a unified framework.
Separately, other methods address the inherent characteristics of the restoration task itself. 
AdaIR~\cite{cui2025adair} adaptively modulates frequency information, while TEAFormer~\cite{hu2025teaformer} focuses on translation equivariance by proposing an adaptive sliding indexing mechanism.
Despite their adaptive structure, they often lack high-level semantic guidance, making it difficult to distinguish complex composite degradations.

\vspace{0.1cm}
\noindent 
\textbf{Prompt-driven Image Restoration.}
To address the need for explicit guidance, prompt-based approaches have been explored~\cite{wang2023promptrestorer, yao2024neural, jiang2024autodir}. 
PromptIR~\cite{potlapalli2023promptir} employs learnable prompts to encode degradation-specific information, which dynamically guide the restoration network. 
InstructIR~\cite{conde2024instructir} replaces learned prompts with language instructions from a text encoder~\cite{reimers2019sentence}, while OneRestore~\cite{guo2024onerestore}  merges versatile scene descriptors, including text embeddings.

To further enhance the prompt, recent methods apply vision-language (VL) models~\cite{radford2021clip, li2022blip, li2023blip2}.
These models learn strong visual-textual alignment and have been widely adopted across tasks such as visual understanding~\cite{wang2022cris, zhong2022regionclip, rao2022denseclip}, image generation~\cite{rombach2022ldm, poole2023dreamfusion, mou2024t2i}, and multi-task learning~\cite{liu2023hierarchical, ren2024adaptive, wang2024m2clip}.
Motivated by this success, VL models—typically CLIP—have been adapted to image restoration~\cite{li2023promptinprompt, duan2024uniprocessor, yang2024language, tian2025dfpir}.
For instance, DA-CLIP~\cite{luo2024daclip} introduces a degradation-aware controller via cross-attention.
MPerceiver~\cite{ai2024multimodal} leverages a CLIP image encoder to adjust multimodal prompts adaptively, while VLUNet~\cite{zeng2025vlunet} aligns features with degradation descriptions, and DFPIR~\cite{tian2025dfpir} guides feature perturbations.
More recently, a separate line of work explores MLLM-based image restoration, leveraging the high-level reasoning and generative capabilities of large-scale MLLMs via agent-based systems~\cite{chen2024restoreagent, zhu2025an} or diffusion-based approaches~\cite{ai2024dreamclear, he2024training_free}.

Despite significant advances, prompt-driven image restoration methods suffer from several limitations.
First, recent prompt-based methods treat degradations as discrete categories, failing to capture the continuous relationships inherent in composite degradations.
Second, recent MLLM-based approaches primarily rely on textual responses generated by the MLLM, overlooking the rich internal visual knowledge encoded within the model.
Our work overcomes these limitations by extending MLLM guidance beyond textual representations, integrating internal visual features via MGFB and leveraging high-level textual semantics via MoFE for expert routing.

% ==================== Methods ====================
\section{Method}
\label{sec:method}
\begin{figure}[t]
  \centering
  \includegraphics[width=\linewidth]{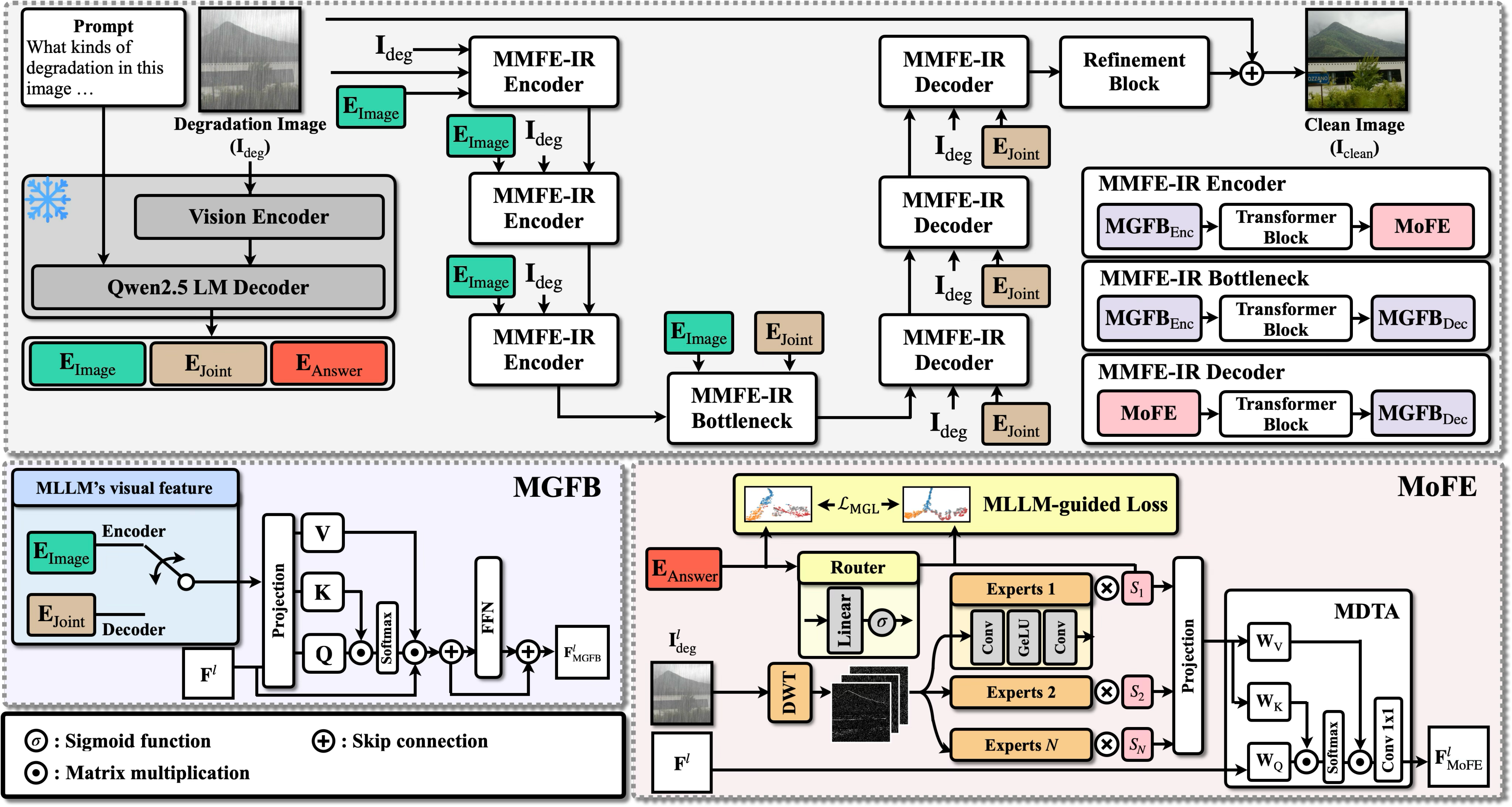}
  \caption{
\textbf{Overview of the proposed MLLM-guided mixture-of-frequency experts for Image Restoration.} Given a degraded image and text prompt, the multimodal large language model (MLLM) extracts embeddings ${\mathbf{E}_{\text{Image}}, \mathbf{E}_{\text{Joint}}, \mathbf{E}_{\text{Answer}}}$ to guide restoration. The MLLM-guided fusion block (MGFB) injects cross-modal information into both encoder and decoder for degradation-aware feature learning. The mixture-of-frequency-experts (MoFE) module further refines representations by dynamically routing frequency components through multiple specialized experts.
  }
  \label{fig:method}
\end{figure}

The overall framework of the proposed MLLM-guided mixture-of-frequency experts for image restoration (MMFE-IR) network is illustrated in \cref{fig:method}.
Given a degraded image $\mathbf{I}_{\rm{deg}}$, the goal is to reconstruct a clean image $\mathbf{I}_{\rm{clean}}$ by leveraging both high-level textual semantics and internal visual knowledge from a pretrained MLLM.

We obtain semantic embeddings $\mathbf{E}_{\mathrm{Image}}$, $\mathbf{E}_{\mathrm{Joint}}$, and $\mathbf{E}_{\mathrm{Answer}}$ by prompting the pretrained MLLM, Qwen2.5-VL~\cite{bai2025qwen2_5},  in a visual question answering (VQA) manner based on its technical report.
This allows the model to generate context-specific representations from the degraded image.
$\mathbf{E}_{\text{Image}}$ represents the contextualized vision-derived embedding pooled from the final decoder's image tokens, $\mathbf{E}_{\text{Joint}}$ denotes vision–language conditioned embeddings pooled from all tokens, and $\mathbf{E}_{\text{Answer}}$ refers to the high-level textual representations of the last token from the language decoder used for text generation.
The restoration backbone integrates these embeddings through an MLLM-guided fusion block (MGFB), and enhances its capability to handle diverse degradations using a mixture-of-frequency-experts (MoFE) module.

\subsection{MLLM-Guided Fusion Block}
\label{sub:MGFB}
Although MLLMs encode rich internal visual knowledge from large-scale pretraining, it is non-trivial to directly utilize this knowledge in low-level image restoration.
Instead, we leverage it as guidance via an MLLM-guided fusion block (MGFB), a cross-attention-based module that assigns MLLM embeddings to different roles and guides the encoder--decoder architecture.

Conventional restoration networks extract general image features in the encoder and perform reconstruction in the decoder~\cite{liang2021swinir, wang2022uformer, zamir2022restormer}.
To enhance the effectiveness of MLLM guidance, we assign $\mathbf{E}_{\rm Image}$ to the encoder and $\mathbf{E}_{\rm Joint}$ to the decoder.
The encoder leverages $\mathbf{E}_{\rm Image}$ to obtain vision-derived representations that strengthen structural feature extraction.
Conversely, the decoder employs $\mathbf{E}_{\rm Joint}$, incorporating vision–language-conditioned representations that facilitate degradation-aware reconstruction.
This complementary assignment integrates structural cues with semantic degradation awareness, leading to more precise restoration.

Given a visual feature map $\mathbf{F}^l$ at the $l$-th layer of the network, we employ a cross-attention mechanism, where the visual feature serves as the query and the MLLM-derived embeddings $\mathbf{E}$ act as the key and value. 
\begin{equation}
\text{Attn}(\mathbf{F}^l, \mathbf{E}) 
= \text{Softmax}\!\left(\frac{\mathbf{QK}^\top}{\sqrt{d}}\right)\mathbf{V},
\end{equation}
where $\mathbf{Q} = \mathbf{W}_Q \mathbf{F}^l$, $\mathbf{K} = \mathbf{W}_K \mathbf{E}$, and $\mathbf{V} = \mathbf{W}_V \mathbf{E}$, with $\mathbf{W}_Q$, $\mathbf{W}_K$, and $\mathbf{W}_V$ denoting learnable projection matrices. 
The fused feature is obtained via a residual connection:
\begin{equation}
\mathbf{F}^{l}_{\rm MGFB} = \mathbf{F}^l + \text{Attn}(\mathbf{F}^l, \mathbf{E}),
\end{equation}
where the MLLM embedding $\mathbf{E}$ is assigned based on the layer's role, such that $\mathbf{E} = \mathbf{E}_{\text{Image}}$ and $\mathbf{E} = \mathbf{E}_{\text{Joint}}$ for the encoder and decoder, respectively.

\subsection{Mixture-of-Frequency Experts}
\label{sub:mofe}
While spatial-domain modeling has been the dominant approach, AdaIR~\cite{cui2025adair} has shown that different degradations exhibit distinct frequency characteristics. 
Inspired by this, we design a mixture-of-frequency-experts (MoFE) module to adaptively capture degradation patterns directly from the original image's frequency domain.

Given an intermediate feature map $\mathbf{F}^l$ at the $l$-th layer, we first interpolate the original degraded image $\mathbf{I}_{\mathrm{deg}}$ to match the resolution of $\mathbf{F}^l$, resulting $\mathbf{I}^l_{\mathrm{deg}}$.
We use this original input rather than $\mathbf{F}^l$ because it contains the undistorted frequency information of the degradation.
$\mathbf{I}^l_{\mathrm{deg}}$ is then decomposed into multiple frequency subbands using the discrete wavelet transform (DWT):
\begin{equation}
\mathbf{F}^l_{\text{freq}} = \text{DWT}(\mathbf{I}_{\mathrm{deg}}^l),
\end{equation}
where $\mathbf{F}^l_{\text{freq}} = \{\mathbf{F}^l_{\text{LL}}, \mathbf{F}^l_{\text{LH}}, \mathbf{F}^l_{\text{HL}}, \mathbf{F}^l_{\text{HH}}\}$ denotes the set of low- and high-frequency components from the original image.

Since self-attention in Transformer inherently captures low-frequency features~\cite{wang2022anti, park2022how}, we exclude the $\mathbf{F}^l_{\text{LL}}$ component and process only the high-frequency subbands $\mathbf{F}^l_{\text{HF}} = \text{Concat}(\mathbf{F}^l_{\text{LH}}, \mathbf{F}^l_{\text{HL}}, \mathbf{F}^l_{\text{HH}})$.
We report an ablation on the effect of excluding $\mathbf{F}^l_{\text{LL}}$ in the supplementary material.
This $\mathbf{F}^l_{\text{HF}}$ is processed by a set of frequency experts $\{\mathcal{M}_i\}_{i=1}^N$, where each expert is a lightweight convolutional block (Conv-GeLU-Conv).
Here, $N$ denotes the number of experts.

The expert weights $\mathbf{S} = \{S_1, S_2, \dots, S_N\}$ are computed by the router, which maps the MLLM's semantic embedding $\mathbf{E}_{\text{Answer}}$ through a linear layer and a sigmoid function $\sigma$ as:
\begin{equation}
\mathbf{S} = \sigma(\mathbf{W}_r \mathbf{E}_{\text{Answer}}).
\label{eq:routing}
\end{equation}
The output of the experts is computed as the weighted sum, $\sum_{i=1}^N S_i \cdot \mathcal{M}_i(\mathbf{F}^l_{\text{HF}})$.
This aggregated feature is then passed through a projection layer, resulting in the final MLLM-guided frequency feature, $\mathbf{F}'_{\text{HF}}$.

We fuse this processed frequency information $\mathbf{F}'_{\text{HF}}$ into $\mathbf{F}^l$ using Multi-Dconv Head Transposed Attention (MDTA) from Restormer~\cite{zamir2022restormer}.
$\mathbf{F}^l$ serves as the query ($\mathbf{Q}$), while the MLLM-guided frequency feature $\mathbf{F}'_{\text{HF}}$ acts as the key ($\mathbf{K}$) and value ($\mathbf{V}$):
\begin{equation}
\mathbf{Q} = \mathbf{W}_Q \mathbf{F}^l, \quad \mathbf{K} = \mathbf{W}_K \mathbf{F}'_{\text{HF}}, \quad \mathbf{V} = \mathbf{W}_V \mathbf{F}'_{\text{HF}}.
\end{equation}
The attention output is passed through a $1 \times 1$ convolutional layer and added to the original feature via a residual connection:
\begin{equation}
\mathbf{F}^{l+1} = \mathbf{F}^l + \text{Conv}_{1 \times 1}\left(\text{Softmax}\left(\frac{\mathbf{QK}^\top}{\sqrt{d}}\right)\mathbf{V}\right).
\label{eq:mofe_fusion}
\end{equation}
This allows the spatial features $\mathbf{F}^l$ to adaptively query and integrate the necessary high-frequency details, guided by the MLLM's semantic understanding.

\subsection{MLLM-guided Router Optimization}
\label{sub:simloss}
The router is trained to preserve the relational structure of degradation representations captured by the MLLM.
As shown in \cref{fig:latent}, the MLLM encodes images into a structured embedding space. In this semantic space, the degradation embeddings are distributed across a continuous spectrum. This means that complex or mixed degradations are positioned semantically between the relevant simple degradation clusters.
To ensure that the router preserves this relationship, we introduce an MLLM-guided loss that aligns the pairwise relations in the MLLM feature space with those in the router outputs.

For each mini-batch, we first compute the pairwise cosine similarity matrix of the high-level MLLM embeddings $\mathbf{E}_{\text{Answer}}$, denoted as $\text{Sim}(\mathbf{E}_{\text{Answer}}) \in \mathbb{R}^{B \times B}$, where $B$ is the batch size.
Here, $\text{Sim}(\cdot)$ computes pairwise cosine similarities between all samples within a batch.
After passing $\mathbf{E}_{\text{Answer}}$ through the router, we obtain expert weights $\mathbf{S} \in \mathbb{R}^{B \times N}$, as defined in \cref{eq:routing}.
The MLLM-guided loss $\mathcal{L}_{\text{MGL}}$ is defined as:
\begin{equation}
\mathcal{L}_{\text{MGL}} = \left\| \text{Sim}(\mathbf{E}_{\text{Answer}}) - \text{Sim}(\mathbf{S}) \right\|_1.
\end{equation}
This loss encourages the router to preserve the relational structure of the MLLM embeddings, where samples with similar degradations produce similar routing scores, while those with different degradations form distinct expert combinations.
As a result, the router effectively leverages the semantically continuous relationships captured by the MLLM.

The total training loss combines this relational consistency with the reconstruction loss:
\begin{equation}
\mathcal{L}_{\text{total}} = \mathcal{L}_{\text{rec}} + \lambda \mathcal{L}_{\text{MGL}},
\label{eq:total_loss}
\end{equation}
where $\mathcal{L}_{\text{rec}}$ is the reconstruction loss, which measures the fidelity between the restored image and the ground-truth image, and $\lambda$ is control parameter.

Since MLLM features are sensitive to input resolution~\cite{bhagwatkar-etal-2024-improving, peng2024qwen2_VL, vasu2025fastvlm}, we further enhance the stability of router optimization through a progressive multi-scale training strategy inspired by Restormer~\cite{zamir2022restormer}.
The strategy gradually increases the input resolution, allowing the router and restoration backbone to adapt smoothly to higher-resolution inputs.
This improves both semantic alignment and restoration robustness across varying image resolutions.

% ==================== Experiments ====================
\section{Experiments}
Following previous all-in-one image restoration works~\cite{guo2024onerestore, cui2025adair, tian2025dfpir, zamfir2025moceir}, we conduct comprehensive experiments under the composite degradation and all-in-one settings.
The composite setting trains the model to handle mixed cases of up to three degradation types, while the all-in-one setting involves joint training on three- and five-degradations.

\subsection{Experimental Settings}
\noindent
\textbf{Datasets.}
We adopt the same dataset configuration as prior all-in-one methods~\cite{cui2025adair, zeng2025vlunet, tian2025dfpir, zamfir2025moceir}.
We use the CDD11 dataset~\cite{guo2024cdd11} for the composite degradation setup.
For the all-in-one setting, we use BSD400~\cite{pablo2011bsd400} and WED~\cite{kede2017wed} with Gaussian noise levels $\sigma \in \{15, 25, 50\}$, and evaluate on BSD68~\cite{martin2001bsd68} for denoising.
We employ Rain100L~\cite{yang2020rain100l}, SOTS~\cite{li2018sot}, GoPro~\cite{nah2017gopro}, and LOL-v1~\cite{wei2018lol} for the tasks of deraining, dehazing, motion deblurring, and low-light enhancement, respectively.
% The CDD11 dataset~\cite{guo2024cdd11} is used for the composite degradation setup.

\begin{table}[t]
\centering
\caption{
\textbf{Comparison to state-of-the-art on CDD11 dataset.} 
PSNR (dB, $\uparrow$) and \colorbox{clblue!50}{SSIM ($\uparrow$)}.
The best and second best results are highlighted in \textbf{bold} and \underline{underline}, respectively.
Our method achieves significant improvements on double (L+H, H+S) and triple (L+H+R, L+H+S) compositions, demonstrating strong robustness to composite degradations.
}
\resizebox{\textwidth}{!}{
\begin{tabular}{lcccccccccccccccccccccccc}
\toprule
Methods & \multicolumn{8}{c}{CDD11-Single}   & \multicolumn{10}{c}{CDD11-Double}   & \multicolumn{4}{c}{CDD11-Triple}  & \multicolumn{2}{c}{\multirow{2}{*}{Average}} \\
\cmidrule(lr){2-9} \cmidrule(lr){10-19} \cmidrule(lr){20-23}
            & \multicolumn{2}{c}{Low (L)} & \multicolumn{2}{c}{Haze (H)} & \multicolumn{2}{c}{Rain (R)} & \multicolumn{2}{c}{Snow (S)} & \multicolumn{2}{c}{L+H} & \multicolumn{2}{c}{L+R} & \multicolumn{2}{c}{L+S} & \multicolumn{2}{c}{H+R} & \multicolumn{2}{c}{H+S} & \multicolumn{2}{c}{L+H+R} & \multicolumn{2}{c}{L+H+S} & \multicolumn{2}{c}{} \\
\midrule
AirNet~\cite{li2022airnet}      & 24.83  & \cc{.778}  & 24.21  & \cc{.951}  & 26.55  & \cc{.891}  & 26.79  & \cc{.919}  & 23.23  & \cc{.779}  & 22.82  & \cc{.710}  & 23.29  & \cc{.723}  & 22.21  & \cc{.868}  & 23.29  & \cc{.901}  & 21.80  & \cc{.708}  & 22.24  & \cc{.725}  & 23.75  & \cc{.814}  \\
PromptIR~\cite{potlapalli2023promptir}    & 26.32  & \cc{.805}  & 26.10  & \cc{.969}  & 31.56  & \cc{.946}  & 31.53  & \cc{.960}  & 24.49  & \cc{.789}  & 25.05  & \cc{.771}  & 24.51  & \cc{.761}  & 24.54  & \cc{.924}  & 23.70  & \cc{.925}  & 23.74  & \cc{.752}  & 23.33  & \cc{.747}  & 25.90  & \cc{.850}  \\
WGWSNet~\cite{zhu2023wgwsnet}     & 24.39  & \cc{.774}  & 27.90  & \cc{.982}  & 33.15  & \cc{.964}  & 34.43  & \cc{.973}  & 24.27  & \cc{.800}  & 25.06  & \cc{.772}  & 24.60  & \cc{.765}  & 27.23  & \cc{.955}  & 27.65  & \cc{.960}  & 23.90  & \cc{.772}  & 23.97  & \cc{.771}  & 26.96  & \cc{.863}  \\
WeatherDiff~\cite{ozdenizci2023weatherdiff} & 23.58  & \cc{.763}  & 21.99  & \cc{.904}  & 24.85  & \cc{.885}  & 24.80  & \cc{.888}  & 21.83  & \cc{.756}  & 22.69  & \cc{.730}  & 22.12  & \cc{.707}  & 21.25  & \cc{.868}  & 21.99  & \cc{.868}  & 21.23  & \cc{.716}  & 21.04  & \cc{.698}  & 22.49  & \cc{.798}  \\
OneRestore~\cite{guo2024onerestore}  & 26.48  & \cc{\underline{.826}}  & 32.52  & \cc{\underline{.990}}  & 33.40  & \cc{.964}  & 34.31  & \cc{.973}  & 25.79  & \cc{\underline{.822}}  & 25.58  & \cc{.799}  & 25.19  & \cc{.789}  & \underline{29.99}  & \cc{.957}  & \underline{30.21}  & \cc{.964}  & 24.78  & \cc{.788}  & 24.90  & \cc{.791}  & 28.47  & \cc{.878}  \\
MoCEIR~\cite{zamfir2025moceir}      & \underline{27.26}  & \cc{.824}  & \underline{32.66}  & \cc{\underline{.990}}  & \underline{34.31}  & \cc{\underline{.970}}  & \underline{35.91}  & \cc{\underline{.980}}  & \underline{26.24}  & \cc{.817}  & \underline{26.25}  & \cc{\underline{.800}}  & \underline{26.04}  & \cc{\underline{.793}}  & 29.93  & \cc{\underline{.964}}  & 30.19  & \cc{\underline{.970}}  & 25.41  & \cc{.789}  & 25.39  & \cc{.790}  & \underline{29.05}  & \cc{\underline{.881}}  \\
TEAFormer~\cite{hu2025teaformer}   & -      & \cc{-   }  & -      & \cc{-   }  & -      & \cc{-   }  & -      & \cc{-   }  & -      & \cc{-   }  & -      & \cc{-   }  & -      & \cc{-   }  & -      & \cc{-   }  & -      & \cc{-   }  & \underline{25.88}  & \cc{\underline{.796}}  & \underline{26.05}  & \cc{\underline{.800}}  & -      & \cc{-   }  \\
\midrule
Ours        & \textbf{27.73}  & \cc{\textbf{.830}}  & \textbf{36.03}  & \cc{\textbf{.993}}  & \textbf{35.44}  & \cc{\textbf{.975}}  & \textbf{37.25}  & \cc{\textbf{.984}}  & \textbf{27.07}  & \cc{\textbf{.826}}  & \textbf{26.85}  & \cc{\textbf{.809}}  & \textbf{26.82}  & \cc{\textbf{.806}}  & \textbf{32.08}  & \cc{\textbf{.972}}  & \textbf{32.72}  & \cc{\textbf{.977}}  & \textbf{26.13}  & \cc{\textbf{.800}}  & \textbf{26.28}  & \cc{\textbf{.802}}  & \textbf{30.40}  & \cc{\textbf{.889}}  \\
\bottomrule
\end{tabular}
}
\vspace{-0.2cm}
\label{tab:cdd11-quantitative}
\end{table}

\begin{figure}[t]
  \centering
  \includegraphics[width=\linewidth]{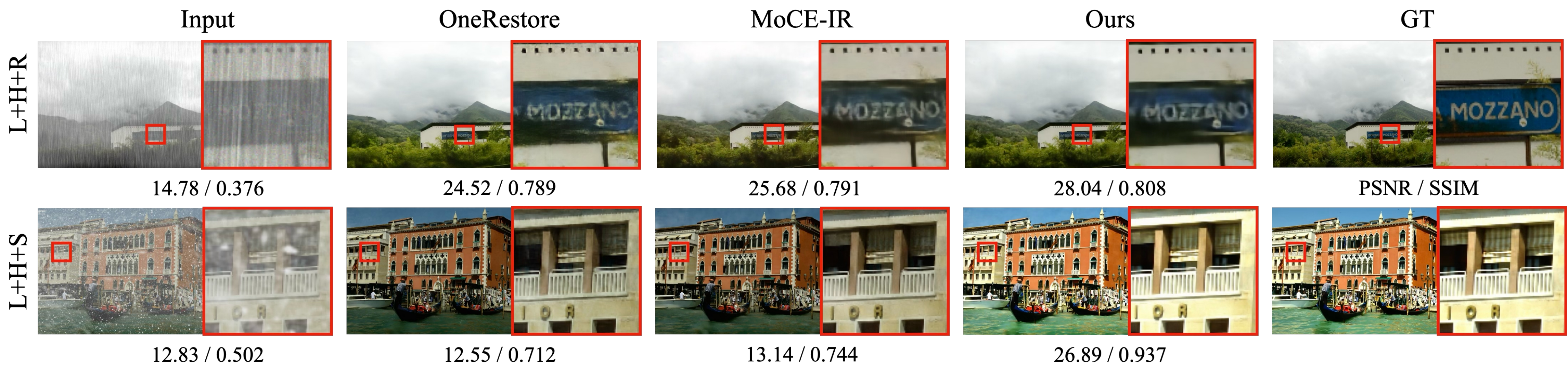}
  \caption{
\textbf{Qualitative results on CDD11 dataset.}
Our method restores clearer textures and natural colors even when multiple degradations coexist, while previous methods tend to leave residual artifacts or lose fine details. 
  }
\vspace{-0.3cm}
  \label{fig:cdd11-qualitative}
\end{figure}

\noindent
\textbf{Implementation Details.}
We adopt Restormer~\cite{zamir2022restormer} as the backbone and Qwen 2.5-VL~\cite{bai2025qwen2_5} as the MLLM.
To obtain the MLLM embeddings, we use VQA prompts (\eg, "What kinds of degradation in this image?") generated via GPT~\cite{achiam2023gpt}.
Our MoFE module uses $N=8$ experts.
The total training loss, \cref{eq:total_loss}, is optimized with $\lambda=0.1$, where $\mathcal{L}_{\text{rec}}$ is the L1 loss in the RGB and Fourier domain, following MoCEIR~\cite{zamfir2025moceir}.
We use the Adam optimizer~\cite{kingma2015adam} ($\beta_1{=}0.9, \beta_2{=}0.999$) with an initial $2\times10^{-4}$ learning rate and cosine decay.
Models are trained for 150 epochs (all-in-one) or 250 epochs (Composite) on 8 NVIDIA RTX A6000 GPUs.
For progressive training, the resolution gradually increases from $128^2$ to $224^2$ while the batch size decreases from 64 to 16. 
More details are provided in the supplementary material.

\begin{table}[t]
\centering
\caption{
\textbf{Comparison to state-of-the-art on five degradations.}
PSNR (dB, $\uparrow$) and \colorbox{clblue!50}{SSIM ($\uparrow$)}.
The best, second, and third best results are highlighted in \textbf{bold}, \underline{underline}, and \emph{italic}, respectively.
The $\dagger$ indicates results from a model trained with the additional Rain13K dataset~\cite{jiang2020multi} to address the data scarcity of the deraining.
}
\vspace{-0.3cm}
\resizebox{0.9\textwidth}{!}{
\begin{tabular}{llcccccccccccc}
\toprule
\multicolumn{2}{c}{Methods} & \multicolumn{2}{c}{\textit{Dehazing}} & \multicolumn{2}{c}{\textit{Deraining}} & \multicolumn{2}{c}{\textit{Denoising}}               & \multicolumn{2}{c}{\textit{Deblurring}} & \multicolumn{2}{c}{\textit{Low-Light}} & \multicolumn{2}{c}{Average}          \\
\cmidrule(lr){3-4} \cmidrule(lr){5-6} \cmidrule(lr){7-8} \cmidrule(lr){9-10}  \cmidrule(lr){11-12}
\multicolumn{2}{c}{}        & \multicolumn{2}{c}{SOTS}              & \multicolumn{2}{c}{Rain 100L}          & \multicolumn{2}{c}{BSD68\textsubscript{$\sigma=25$}} & \multicolumn{2}{c}{GoPro}               & \multicolumn{2}{c}{LoL}                &  \multicolumn{2}{c}{}\\
\midrule
AirNet~\cite{li2022airnet}       & CVPR 2022  & 21.04             & \cc{0.884}             & 32.98             & \cc{0.951}             & 30.91             & \cc{0.882}             & 24.35             & \cc{0.781}             & 18.18             & \cc{0.735}             & 25.49             & \cc{0.847} \\
IDR~\cite{zhang2023idr}          & CVPR 2023  & 25.24             & \cc{0.943}             & 35.63             & \cc{0.965}             & \textbf{31.60}    & \cc{0.887}             & 27.87             & \cc{0.846}             & 21.34             & \cc{0.893}             & 28.34             & \cc{0.907} \\
PromptIR~\cite{potlapalli2023promptir}     & NeurIPS 2023  & 26.54             & \cc{0.949}             & 36.37             & \cc{0.970}             & 31.47             & \cc{0.886}             & 28.71             & \cc{0.881}             & 22.68             & \cc{0.832}             & 29.15             & \cc{0.904} \\
GridFormer~\cite{wang2024gridformer}   & IJCV 2024  & 26.79             & \cc{0.951}             & 36.61             & \cc{0.971}             & 31.45             & \cc{0.885}             & 29.22             & \cc{0.884}             & 22.59             & \cc{0.831}             & 29.33             & \cc{0.904} \\
InstructIR~\cite{conde2024instructir}   & ECCV 2024  & 27.10             & \cc{0.956}             & 36.84             & \cc{0.973}             & 31.40             & \cc{0.887}             & 29.40             & \cc{0.886}             & 23.00             & \cc{0.836}             & 29.55             & \cc{0.908} \\
VLU-Net~\cite{zeng2025vlunet}      & CVPR 2025  & 30.84             & \cc{\underline{0.980}} & \underline{38.54} & \cc{\underline{0.982}} & 31.43             & \cc{\emph{0.891}} & 27.46             & \cc{0.840}             & 22.29             & \cc{0.833}             & 30.11             & \cc{0.905} \\
AdaIR~\cite{cui2025adair}        & ICLR 2025  & 30.53             & \cc{0.978}             & 38.02             & \cc{\emph{0.981}}             & 31.35             & \cc{0.889}             & 28.12             & \cc{0.858}             & 23.00             & \cc{0.845}             & 30.20             & \cc{0.910} \\
MoCE-IR~\cite{zamfir2025moceir}      & CVPR 2025  & 30.48             & \cc{0.974}             & \emph{38.04}             & \cc{\underline{0.982}} & 31.34             & \cc{0.887}             & 30.05             & \cc{\emph{0.899}}             & 23.00             & \cc{0.852}             & 30.58             & \cc{\emph{0.919}} \\
DFPIR~\cite{tian2025dfpir}        & CVPR 2025  & \emph{31.64} & \cc{\emph{0.979}}             & 37.62             & \cc{0.978}             & 31.29             & \cc{0.889}             & 28.82             & \cc{0.873}             & \textbf{23.82}    & \cc{0.843}             & 30.64             & \cc{0.912} \\
TEAFormer~\cite{hu2025teaformer}    & ICCV 2025  & 31.57             & \cc{\underline{0.980}} & \textbf{39.44}    & \cc{\textbf{0.986}}    & 31.54             & \cc{0.890}             & \emph{30.53} & \cc{\underline{0.908}} & 23.06             & \cc{\emph{0.856}} & \emph{31.23} & \cc{\underline{0.924}} \\
\midrule
Ours         &         & \underline{32.16} & \cc{\textbf{0.981}}    & 37.86             & \cc{\underline{0.982}} & \emph{31.56} & \cc{\underline{0.895}}    & \textbf{32.23}    & \cc{\textbf{0.932}}    & \underline{23.40} & \cc{\textbf{0.866}}    & \underline{31.44}    & \cc{\textbf{0.931}} \\
Ours$\dagger$&         & \textbf{32.24}    & \cc{\textbf{0.981}}    & \underline{38.54} & \cc{\underline{0.982}} & \underline{31.59} & \cc{\textbf{0.896}}    & \underline{32.22}    & \cc{\textbf{0.932}}    & \emph{23.18} & \cc{\underline{0.862}}    & \textbf{31.55}    & \cc{\textbf{0.931}} \\
\bottomrule
\end{tabular}
\vspace{-0.3cm}
}
\label{tab:5-quantitative}
\end{table}

\begin{table}[t]
\centering
\caption{
\textbf{Comparison to state-of-the-art on three degradations.} 
PSNR (dB, $\uparrow$) and \colorbox{clblue!50}{SSIM ($\uparrow$)}.
The best and second best results are highlighted in \textbf{bold} and \underline{underline}, respectively.
Our method achieves the highest overall performance, particularly excelling in dehazing and denoising, which involve both low- and high-frequency restoration.
}
\vspace{-0.3cm}
\resizebox{0.9\textwidth}{!}{
\begin{tabular}{llcccccccccccc}
\toprule
\multicolumn{2}{c}{Methods} & \multicolumn{2}{c}{\textit{Dehazing}} & \multicolumn{2}{c}{\textit{Deraining}} & \multicolumn{6}{c}{\textit{Denoising}} & \multicolumn{2}{c}{Average} \\
\cmidrule(lr){3-4} \cmidrule(lr){5-6} \cmidrule(lr){7-12}
\multicolumn{2}{c}{}                         & \multicolumn{2}{c}{SOTS}             & \multicolumn{2}{c}{Rain 100L}                & \multicolumn{2}{c}{BSD68\textsubscript{$\sigma=15$}} & \multicolumn{2}{c}{BSD68\textsubscript{$\sigma=25$}} & \multicolumn{2}{c}{BSD68\textsubscript{$\sigma=50$}} & \multicolumn{2}{c}{}  \\
\midrule
AirNet~\cite{li2022airnet}     & CVPR 2022 & 27.94             & \cc{0.962}             & 34.90             & \cc{0.967}            & 33.92             & \cc{0.933}             & 31.26              & \cc{0.888}             & 28.00             & \cc{0.797}             & 31.20             & \cc{0.909}        \\
IDR~\cite{zhang2023idr}        & CVPR 2023 & 29.87             & \cc{0.970}             & 36.03             & \cc{0.971}            & 33.89             & \cc{0.931}             & 31.32              & \cc{0.884}             & 28.04             & \cc{0.798}             & 31.83             & \cc{0.911}        \\
PromptIR~\cite{potlapalli2023promptir}   & NeurIPS 2023 & 30.58             & \cc{0.974}             & 36.37             & \cc{0.972}            & 33.98             & \cc{0.933}             & 31.31              & \cc{0.888}             & 28.06             & \cc{0.799}             & 32.06             & \cc{0.913}        \\
GridFormer~\cite{wang2024gridformer} & IJCV 2024 & 30.37             & \cc{0.970}             & 37.15             & \cc{0.972}            & 33.93             & \cc{0.931}             & 31.37              & \cc{0.887}             & 28.11             & \cc{0.801}             & 32.19             & \cc{0.912}        \\
InstructIR~\cite{conde2024instructir} & ECCV 2024 & 30.22             & \cc{0.959}             & 37.98             & \cc{0.978}            & \underline{34.15} & \cc{0.933}             & \underline{31.52}  & \cc{0.890}             & \underline{28.30} & \cc{0.804}             & 32.43             & \cc{0.913}        \\
AdaIR~\cite{cui2025adair}      & ICLR 2025 & 31.06             & \cc{\underline{0.980}} & 38.64             & \cc{\underline{0.983}}& 34.12             & \cc{\underline{0.935}} & 31.45              & \cc{0.892}             & 28.19             & \cc{0.802}             & 32.69             & \cc{0.918}        \\
VLU-Net~\cite{zeng2025vlunet}    & CVPR 2025 & 30.71             & \cc{\underline{0.980}} & \textbf{38.93}    & \cc{\textbf{0.984}}   & 34.13             & \cc{\underline{0.935}} & 31.48              & \cc{0.892}             & 28.23             & \cc{0.804}             & 32.70             & \cc{\underline{0.919}}        \\
MoCE-IR~\cite{zamfir2025moceir}    & CVPR 2025 & 31.34             & \cc{0.979}             & 38.57             & \cc{\textbf{0.984}}   & 34.11             & \cc{0.932}             & 31.45              & \cc{0.888}             & 28.18             & \cc{0.800}             & 32.73             & \cc{0.917}        \\
DFPIR~\cite{tian2025dfpir}      & CVPR 2025 & \underline{31.87} & \cc{\underline{0.980}} & \underline{38.65} & \cc{0.982}            & 34.14             & \cc{\underline{0.935}} & 31.47              & \cc{\underline{0.893}} & 28.25             & \cc{\underline{0.806}} & \underline{32.88} & \cc{\underline{0.919}}        \\
\midrule
Ours      &        & \textbf{32.13}     & \cc{\textbf{0.982}}    & 38.34             & \cc{\underline{0.983}}& \textbf{34.23}    & \cc{\textbf{0.938}}    & \textbf{31.61}     & \cc{\textbf{0.896}}    & \textbf{28.39}    & \cc{\textbf{0.812}}    & \textbf{32.94}    & \cc{\textbf{0.922}}        \\
\bottomrule
\end{tabular}
\vspace{-0.3cm}
}
\label{tab:3-quantitative}
\end{table}

\begin{figure}[ht]
  \centering
  \includegraphics[width=\linewidth]{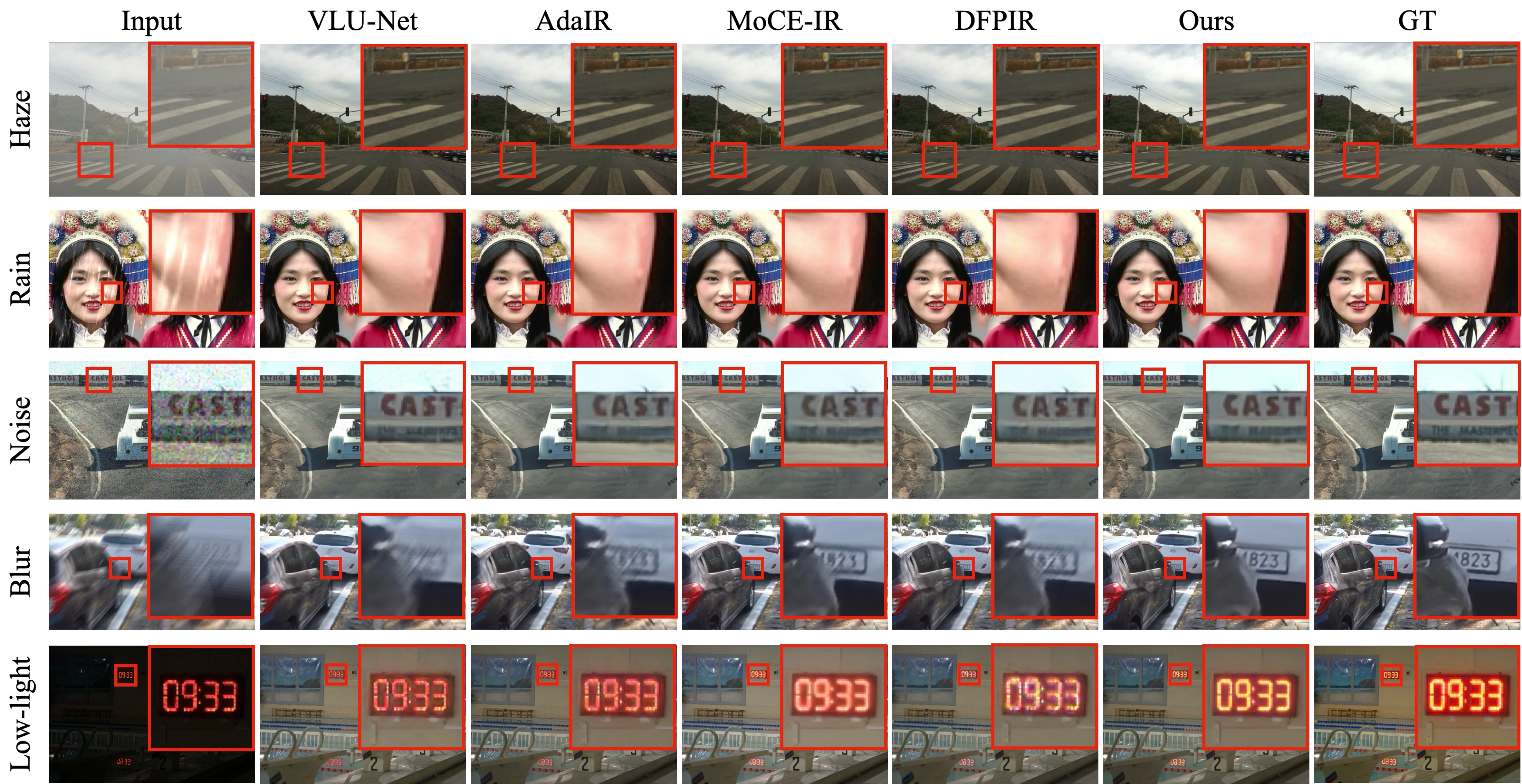}
  \caption{
\textbf{Qualitative results on five degradation.}
The proposed method restores clearer structures, sharper textures, and more natural color tones compared to competing approaches, which often produce over-smoothed or artifact-prone results.
}
\vspace{-0.3cm}
  \label{fig:5d-qualitative}
\end{figure}

\subsection{Comparison with State-of-the Art Methods}
We compare our method against recent state-of-the-art (SOTA) approaches. 
We lead with the composite degradation benchmark, which best demonstrates our model's robustness, followed by the standard five- and three-degradation all-in-one settings.

\noindent \textbf{Composite Degradation Results.} 
As shown in \cref{tab:cdd11-quantitative}, our method demonstrates exceptional robustness on the challenging CDD11 dataset, achieving the highest average PSNR (30.40 dB) and SSIM (0.889), and outperforming all prior approaches by a large margin. 
Unlike competing methods that degrade significantly when multiple degradations coexist, our method maintains strong performance not only on single degradations, but also on challenging double (L+H, H+S) and triple (L+H+R, L+H+S) compositions. 
\cref{fig:cdd11-qualitative} presents qualitative comparisons under the challenging three composite degradation scenarios.
Our method produces visually faithful results, effectively restoring fine textures and realistic color tones even when multiple degradations occur simultaneously.
In contrast, competing approaches tend to leave residual artifacts, particularly in regions affected by overlapping degradations.
These results highlight the robustness of our method in handling complex degradation compositions.
Further qualitative results are available in the supplementary material.

\begin{table}[t]
\centering
\caption{
\textbf{Ablation studies.}
(a) Ablation of MGFB input types and MoFE.
Using $\mathbf{E}_{\text{Image}}$ for the encoder and $\mathbf{E}_{\text{Joint}}$ for the decoder yields the best results, while MoFE further enhances performance efficiently.
(b) Quantitative ablation on MLLM-guided loss and progressive multi-scale training.
Both components consistently improve restoration performance, and their combination achieves the highest PSNR and SSIM.
}
\vspace{-0.5cm}
\label{tab:ablation_combined}

% -------------------- (a) --------------------
\begin{subtable}[t]{0.58\textwidth}
\centering
\caption{MGFB input types and MoFE.}
\label{tab:abl-input}
\resizebox{\linewidth}{!}{
\begin{tabular}{cc|cc|c|c|cc}
\toprule
\multicolumn{2}{c|}{MGFB\textsubscript{Enc}} &
\multicolumn{2}{c|}{MGFB\textsubscript{Dec}} &
\multirow{2}{*}{MoFE} &
\multirow{2}{*}{\# Param (M)} &
\multicolumn{2}{c}{Average} \\
\cmidrule(lr){1-2} \cmidrule(lr){3-4} \cmidrule(lr){7-8}
$\mathbf{E}_{\rm Image}$ & $\mathbf{E}_{\rm Joint}$ &
$\mathbf{E}_{\rm Image}$ & $\mathbf{E}_{\rm Joint}$ &
& & PSNR & SSIM \\
\midrule
\checkmark &  &  & \checkmark & \checkmark & 37.5 & 30.40 & .889 \\
\midrule
\checkmark &  &  &  &  & \multirow{2}{*}{30.0} & 29.84 & .886 \\
 & \checkmark &  &  &  &  & 29.67 & .885 \\
\midrule
 &  & \checkmark &  &  & \multirow{2}{*}{30.2} & 29.50 & .884 \\
 &  &  & \checkmark &  &  & 29.72 & .887 \\
\midrule
 &  &  &  & \checkmark & 29.6 & 29.86 & .887 \\
\midrule
 &  &  &  &  & 26.1 & 28.84 & .881 \\
\bottomrule
\end{tabular}
}
\end{subtable}
\hfill
% -------------------- (b) --------------------
\begin{subtable}[t]{0.38\textwidth}
\centering
\caption{MLLM-guided loss and progressive multi-scale training.}
\label{tab:abl-training}
\resizebox{\linewidth}{!}{
\begin{tabular}{cccc}
\toprule
\multirow{2}{*}{\begin{tabular}[c]{@{}c@{}}MLLM-guided\\ Loss\end{tabular}} &
\multirow{2}{*}{\begin{tabular}[c]{@{}c@{}}Multi-scale\\ Training\end{tabular}} &
\multicolumn{2}{c}{Average} \\
\cmidrule(lr){3-4}
& & PSNR & SSIM \\
\midrule
\checkmark & \checkmark & 30.40 & 0.889 \\
 & \checkmark & 30.10 & 0.888 \\
 &  & 29.72 & 0.885 \\
\bottomrule
\end{tabular}
}
\end{subtable}
\end{table}

\begin{table}[t]
\centering
\caption{
\textbf{Analysis of expert scaling, domain design, and guide models.}
(a) Impact of the number of experts.
Increasing the number of experts improves performance with marginal parameter growth.
(b) Comparison of MoE domain design.
Frequency-domain MoE outperforms spatial-domain MoE.
(c) Comparison of different guide models.
MLLM guidance yields superior performance compared to prompt-based guidance.
}
\vspace{-0.5cm}
\label{tab:expert_analysis}
% ---------------- (a) ----------------
\begin{subtable}[t]{0.28\linewidth}
\centering
\caption{Number of experts.}
\label{tab:abl-expert-scaling}
\resizebox{\linewidth}{!}{
\begin{tabular}{c|cc|c}
\toprule
\# Experts & PSNR & SSIM & \# Param (M) \\
\midrule
1  & 30.04 & 0.888 & 35.9 \\
2  & 30.09 & 0.888 & 36.1 \\
4  & 30.33 & 0.889 & 36.6 \\
\textbf{8}  & \textbf{30.40} & \textbf{0.889} & \textbf{37.5} \\
16 & 30.43 & 0.889 & 39.3 \\
\bottomrule
\end{tabular}
}
\end{subtable}
\hfill
% ---------------- (b) ----------------
\begin{subtable}[t]{0.23\linewidth}
\centering
\caption{MoE domain.}
\label{tab:abl-domain}
\resizebox{\linewidth}{!}{
\begin{tabular}{c|cc}
\toprule
MoE Domain & PSNR & SSIM \\
\midrule
Frequency & 30.40 & 0.889 \\
Spatial   & 30.16 & 0.888 \\
\bottomrule
\end{tabular}
}
\end{subtable}
\hfill
% ---------------- (c) ----------------
\begin{subtable}[t]{0.47\linewidth}
\centering
\caption{Guide model comparison.}
\label{tab:abl-guidemodel}
\resizebox{\linewidth}{!}{
\begin{tabular}{ll|cc|l}
\toprule
\multicolumn{2}{c|}{Guide Model} & PSNR & SSIM & Latency (s)\\
\midrule
\multirow{3}{*}{MLLM}
  & Qwen2.5-VL 3B & 30.40 & 0.889 & 1.75 (\textcolor{blue}{+67\%}) \\
  & FastVLM-1.5B  & 30.50 & 0.890 & 1.15 (\textcolor{blue}{+10\%}) \\
  & FastVLM-0.5B  & 30.45 & 0.889 & 1.12 (\textcolor{blue}{+7\%}) \\
\midrule
Prompt-based & OneRestore & 29.73 & 0.885 & 1.05 (+0\%) \\
\bottomrule
\end{tabular}
}
\end{subtable}
\end{table}

\vspace{0.2cm}
\noindent \textbf{All-in-one Results.} 
Our model also achieves SOTA performance in the standard all-in-one benchmarks. 
For the five-degradation setting (\cref{tab:5-quantitative}), our method achieves the highest average PSNR (31.44 dB). 
Notably, our model's strength is highlighted by its exceptional performance on both low-frequency tasks like dehazing (+0.52 dB) and high-frequency tasks like deblurring (+1.70 dB), demonstrating our MoFE's ability to effectively model diverse frequency characteristics. 
This visual superiority is illustrated in \cref{fig:5d-qualitative}. 
While we observe a performance trade-off on the standard deraining benchmark—likely due to its limited dataset size and insufficient hyperparameter tuning—we demonstrate that this performance can be improved with additional data.
This strength in handling diverse tasks is further reflected in the three-degradation benchmark (\cref{tab:3-quantitative}). 
Our model not only achieves the best average PSNR (32.94 dB) but also attains top performance in dehazing and across all noise levels in denoising.

\subsection{Ablation studies}
\noindent
\textbf{Analysis of MGFB.}
We adopt Restormer~\cite{zamir2022restormer} as the baseline and investigate the role of MLLM guidance by analyzing different configurations of MGFB, as shown in \cref{tab:abl-input}.
When the MGFB is applied to the encoder, using the $\mathbf{E}_{\text{Image}}$ achieves higher performance (29.84 dB) than $\mathbf{E}_{\text{Joint}}$ (29.67 dB).
This is because the encoder primarily captures general visual structures and textures, thus benefiting more from the vision-derived features in $\mathbf{E}_{\text{Image}}$.
In contrast, applying $\mathbf{E}_{\text{Joint}}$ to the decoder leads to better performance (29.72 dB) than $\mathbf{E}_{\text{Image}}$ (29.50 dB), as the decoder performs degradation-aware reconstruction and effectively utilizes the vision-language conditioned semantics from $\mathbf{E}_{\text{Joint}}$.

\begin{figure}[t]
  \centering
  \includegraphics[width=0.8\linewidth]{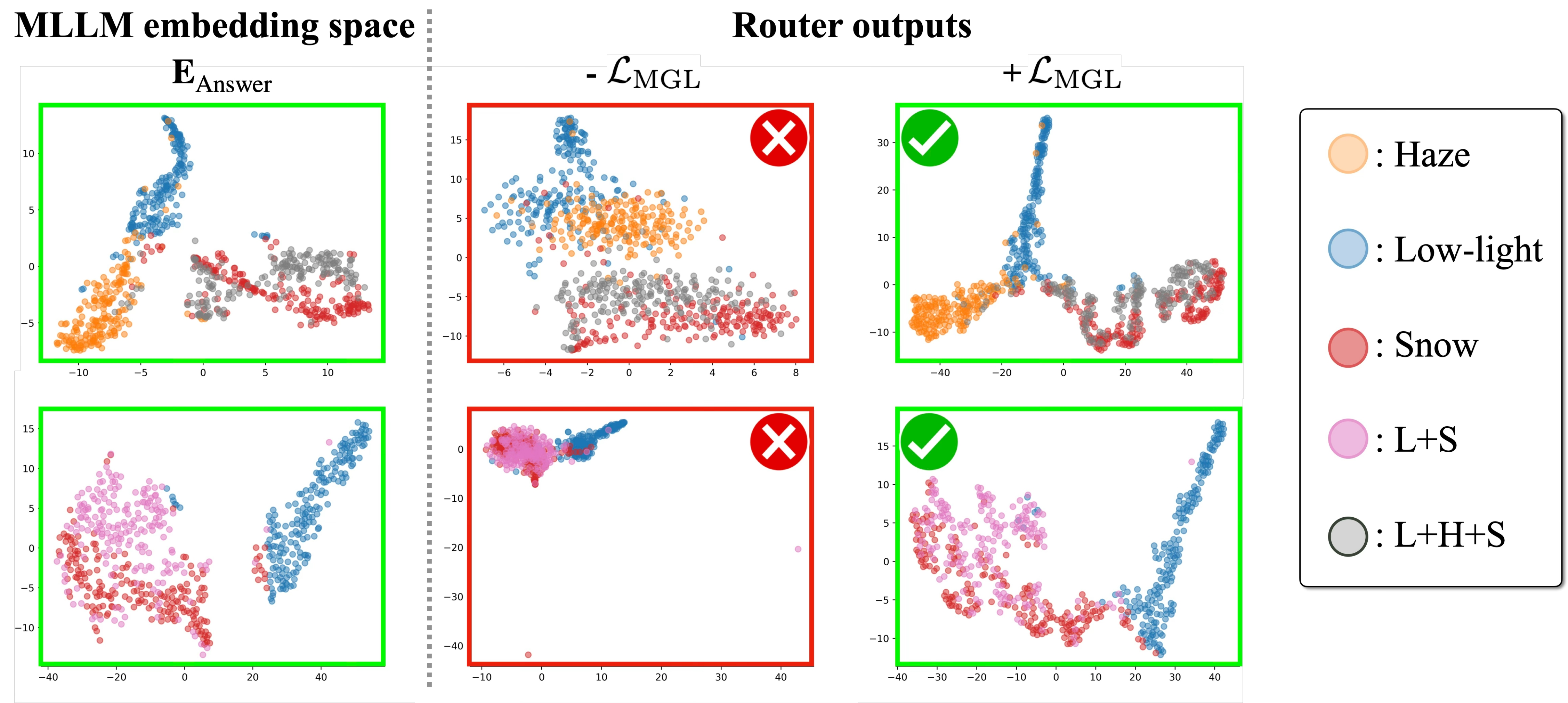}
  \vspace{-0.3cm}
  \caption{
\textbf{
Visualization of degradation feature distributions.} The MLLM-guided loss ($\mathcal L_{\rm MGL}$) encourages the router outputs to align with the MLLM embedding space.
  }
  \label{fig:abl-simloss}
      \vspace{-0.4cm}
\end{figure}

\vspace{0.2cm}
\noindent
\textbf{Effect of MLLM-guided Loss and Multi-scale Training.}
We further analyze the contribution of the MLLM-guided loss and progressive multi-scale training, as summarized in \cref{tab:abl-training}.
First, incorporating progressive multi-scale training improves the overall performance, achieving a gain of +0.38 dB in PSNR.
This improvement can be attributed to the network’s enhanced ability to handle features from various resolutions. As a result, the model not only better utilizes semantic knowledge in MGFB but also improves frequency-aware expert routing in MoFE.
When the MLLM-guided loss is applied, the model exhibits a further improvement in PSNR of +0.3 dB.
As shown in \cref{fig:abl-simloss}, the router outputs become more consistent with the MLLM feature after applying the MLLM-guided loss, where degradation clusters are separated while mixed degradations lie in smooth transitional regions.
This indicates that the MLLM-guided loss effectively maintains semantic continuity between degradation types.

\begin{figure}[t]
  \centering
  \includegraphics[width=\linewidth]{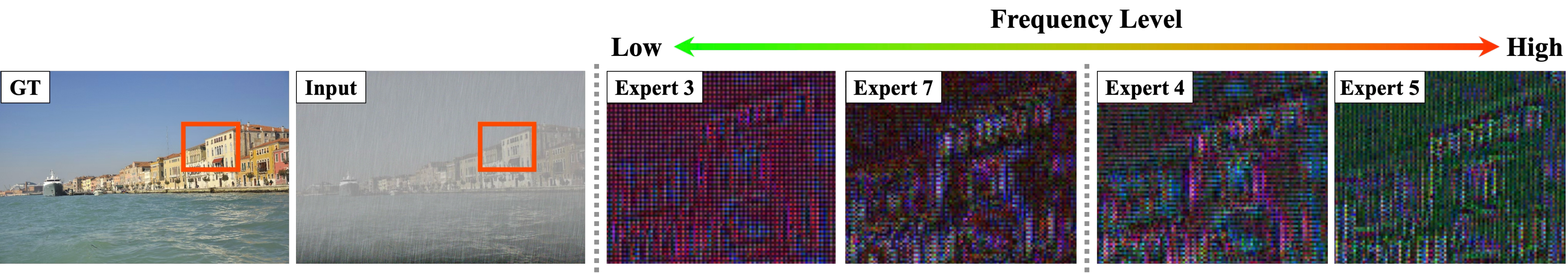}
  \vspace{-0.4cm}
  \caption{
\textbf{Visualization of MoFE's expert difference maps on a composite degraded image (L+H+R).}
Each expert focuses on different frequency levels and characteristics.
The experts specialize in correcting distinct high-frequency components, ranging from Textural Experts (\eg, 3, 7) to Structural Experts (\eg, 4, 5).
  }
    \vspace{-0.4cm}
  \label{fig:abl-expert}
\end{figure}

\vspace{0.1cm}
\noindent
\textbf{Effect of MoFE Module.}
As shown in \cref{tab:abl-input}, incorporating the MoFE module significantly boosts performance, improving the PSNR by +1.02 dB (from 28.84 dB to 29.86 dB) over the baseline, while efficiently adding only 3.5M parameters.
\cref{fig:abl-expert} visualizes the difference maps between the input and the output of each expert.
As our method excludes the low-frequency ($\mathbf{F}^l_{\text{LL}}$) component (\cref{sub:mofe}), all experts specialize in decomposing the relatively high-frequency (HF) sub-bands.
The visualization, which is arranged by increasing frequency level (\cref{fig:abl-expert}), demonstrates that MoFE effectively handles diverse frequency levels.
For example, experts 3 and 7 handle relatively lower frequency levels, considering broader textures and patterns. 
In contrast, experts 4 and 5 handle higher frequency levels by focusing on fine-grained edges or sharp directional structures.
This demonstrates that the MoFE module successfully decomposes the complex restoration task into complementary subspaces by explicitly handling diverse frequency characteristics.

\begin{figure}[t]
  \centering
  \includegraphics[width=\linewidth]{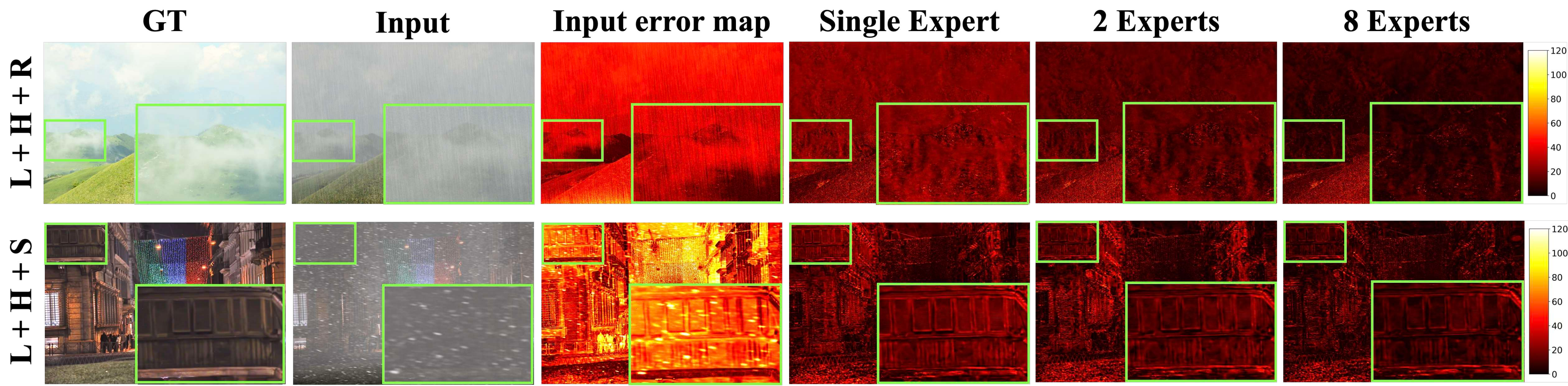}
  \vspace{-0.5cm}
  \caption{
\textbf{Scaling the Number of Experts.}
Increasing the number of experts enables modeling of more diverse degradation characteristics.
}
 \vspace{-0.3cm}
  \label{fig:abl-expert-scaling}
\end{figure}

\begin{figure}[t]
  \centering
  \includegraphics[width=\linewidth]{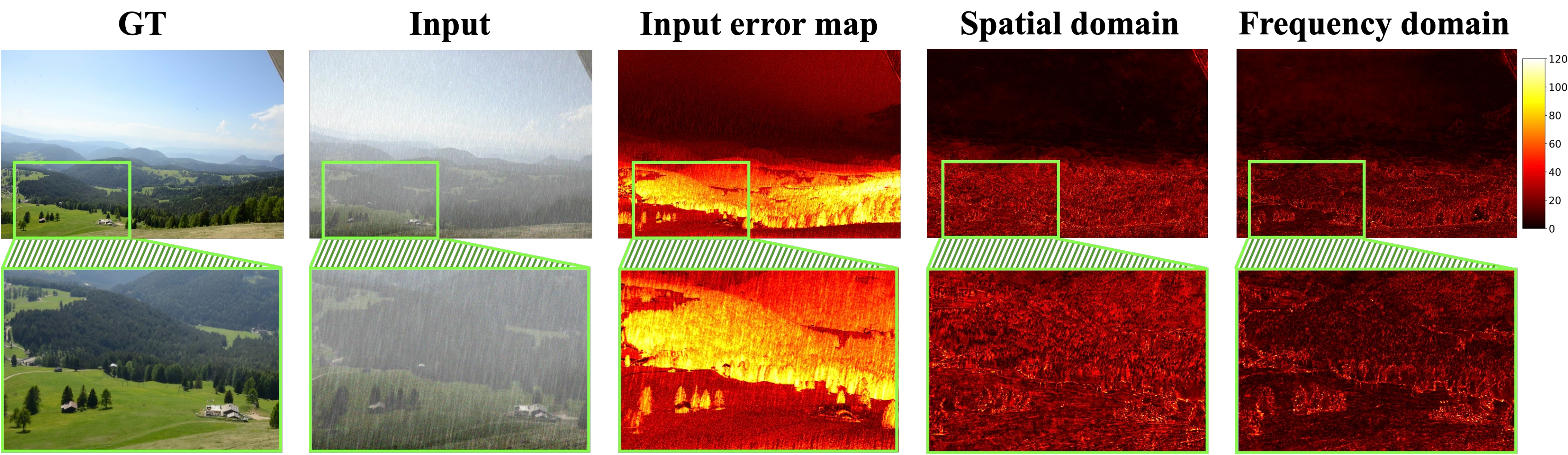}
  \vspace{-0.5cm}
  \caption{
\textbf{MoE in Spatial vs. Frequency Domains.}
MoE in spatial-domain entangles diverse characteristics, making mixed features difficult to disentangle, whereas MoE in frequency-domain separates them across frequency bands for more effective modeling.}
  \vspace{-0.3cm}
  \label{fig:abl-domin}
\end{figure}

\vspace{0.1cm}
\noindent
\textbf{Impact of Expert Scaling.}
\cref{tab:abl-expert-scaling} and \cref{fig:abl-expert-scaling} further analyze the effect of increasing the number of experts.
Performance improves steadily as the expert pool expands, while parameter growth remains marginal.
This indicates that a larger set of experts enhances the capacity to model diverse degradation characteristics.
Notably, performance saturates beyond eight experts, suggesting that sufficient frequency diversity can be captured with a moderate number of specialized experts, achieving a balance between capacity and efficiency.

\vspace{0.1cm}
\noindent
\textbf{Domain Design.}
\cref{tab:abl-domain} and \cref{fig:abl-domin} compares MoE in spatial- and frequency-domain.
Spatial-domain modeling entangles heterogeneous degradations within a shared representation space, making mixed characteristics harder to disentangle.
In contrast, frequency-domain decomposition explicitly separates features across bands, leading to more effective modeling and improved performance.

\vspace{0.1cm}
\noindent
\textbf{Impact of Guide Model.}
\cref{tab:abl-guidemodel} additionally evaluates different guide models.
MLLM-based guidance consistently outperforms prompt-based guidance, indicating that high-level multimodal representations provide more informative and robust signals for expert routing and restoration.
Notably, the performance gap between larger (Qwen2.5-VL-3B~\cite{bai2025qwen2_5}) and smaller (FastVLM-0.5B~\cite{vasu2025fastvlm}) MLLMs is marginal, demonstrating that our framework remains robust across model scales.
Moreover, using a small MLLM incurs only a minor latency overhead compared to prompt-based guidance (1.12s vs. 1.05s), while providing clear gains in restoration quality.
This suggests that lightweight MLLMs are sufficient to provide effective guidance, enabling practical deployment without relying on large-scale models.

% ==================== Conclusion ====================
\section{Conclusion}
We have presented an MLLM-guided all-in-one image restoration framework that overcomes the limitations in the existing methods and successfully leverages high-level semantic knowledge for low-level vision tasks.
Our network incorporates the MGFB to inject vision-derived and language-conditioned representation into the encoder-decoder structure, and an MoFE module to handle diverse frequency-dependent characteristics.
An MLLM-guided router optimization ($\mathcal{L}_{\text{MGL}}$) is introduced to preserve the continuous semantic relationships among degradations, while progressive multi-scale training further enhances stability.
Extensive experiments demonstrate that our method achieves state-of-the-art performance on various benchmarks. 
Notably, our framework shows superior robustness on the challenging CDD11 dataset, proving its effectiveness in handling complex, composite degradations.
% \clearpage  % TODO FINAL: This \clearpage needs to be removed from both review and camera-ready versions.

% \section*{Acknowledgements}
% Please insert your acknowledgments here.

\bibliographystyle{splncs04}
\bibliography{main}

\clearpage
\appendix
\setcounter{figure}{0}
\setcounter{table}{0}
\renewcommand{\thefigure}{S\arabic{figure}}
\renewcommand{\thetable}{S\arabic{table}}
\renewcommand{\thesection}{\Alph{section}}

\vspace*{1em}
\begin{center}
    {\Large \bfseries Supplemental Material}
\end{center}
\vspace{1em}

\section{Detailed Experimental Settings}
\label{sup:detailed_setting}
\subsection{Network Architecture}
We adopt Restormer~\cite{zamir2022restormer} as our backbone architecture, where the number of Transformer blocks from level-1 to level-4 is configured as $[4, 6, 6, 8]$, respectively.
The multi-dconv transposed attention employs attention heads of $[1, 2, 4, 8]$ with corresponding channel dimensions of $[48, 96, 192, 384]$.
In the refinement stage, we utilize 4 blocks, and the gated-dconv feed-forward network operates with a channel expansion factor of $\gamma=2.66$.
The MoFE module is integrated after the downsampling operations and before the upsampling layers.
Additionally, the MGFB is positioned before the Transformer blocks in the encoder stages and after the Transformer blocks in the decoder stages.
To guide the restoration effectively, we integrate the Qwen 2.5-VL~\cite{bai2025qwen2_5} as the MLLM component and employ $N=8$ experts.

\subsection{Training Strategy}
The model is optimized using the Adam optimizer with $\beta_1=0.9$ and $\beta_2=0.999$, initialized with a learning rate of $2 \times 10^{-4}$.
To ensure training stability, we implement a progressive multi-scale training strategy where the input resolution increases while the batch size decreases.
The all-in-one setting spans 150 epochs and proceeds through four stages. 
Specifically, epochs 0--70 use a $128^2$ resolution with a batch size of 64, followed by epochs 70--120 at $160^2$ with batch size 32. 
Subsequently, epochs 120--140 operate at $192^2$ with batch size 16, and the final fine-tuning from epochs 140--150 is performed at $224^2$ with batch size 16.
For the composite setting, the schedule extends to 250 epochs. 
The training starts with epochs 0--130 at $128^2$ (batch 64), transitions to epochs 130--190 at $160^2$ (batch 32), moves to epochs 190--230 at $192^2$ (batch 16), and concludes with epochs 230--250 at $224^2$ (batch 16).

\subsection{Dataset Details}
We follow the dataset configurations of prior all-in-one restoration methods such as PromptIR~\cite{potlapalli2023promptir} and MoCE-IR~\cite{zamfir2025moceir}.
In the case of image denoising, we construct a training set of 5,144 images by combining BSD400 (400 images) and WED (4,744 images), while evaluations are performed on the BSD68 dataset (68 images).
The deraining task utilizes the Rain100L dataset, comprising 200 training and 100 testing pairs.
Regarding dehazing, we adopt the RESIDE-OTS dataset~\cite{li2018sot} with 72,135 training images and test on the 500 pairs from the SOTS outdoor set.
We employ the GoPro dataset (2,103 training and 1,111 testing pairs) for deblurring, and the LOL-v1 dataset (485 training and 15 testing pairs) for low-light enhancement.
Finally, composite degradation experiments are conducted on the CDD11 dataset~\cite{guo2024cdd11}, which is constructed by applying 11 distinct types of composite degradations to 1,383 clear background images.

\section{Additional Experiments}
Due to space constraints in the main manuscript, we focused on the most challenging triple degradation scenarios.
In this section, to demonstrate the robustness of our method across varying degradation complexities, we provide additional qualitative comparisons for single and double degradation combinations.
These results show that our model handles single and double degradations as effectively as complex composite ones.
We also include an additional ablation study on the effect of including the low-frequency component in MoFE.

\subsection{Additional Qualitative Results on Single Degradation}
\begin{figure}[t]
    \centering
    \includegraphics[width=\linewidth]{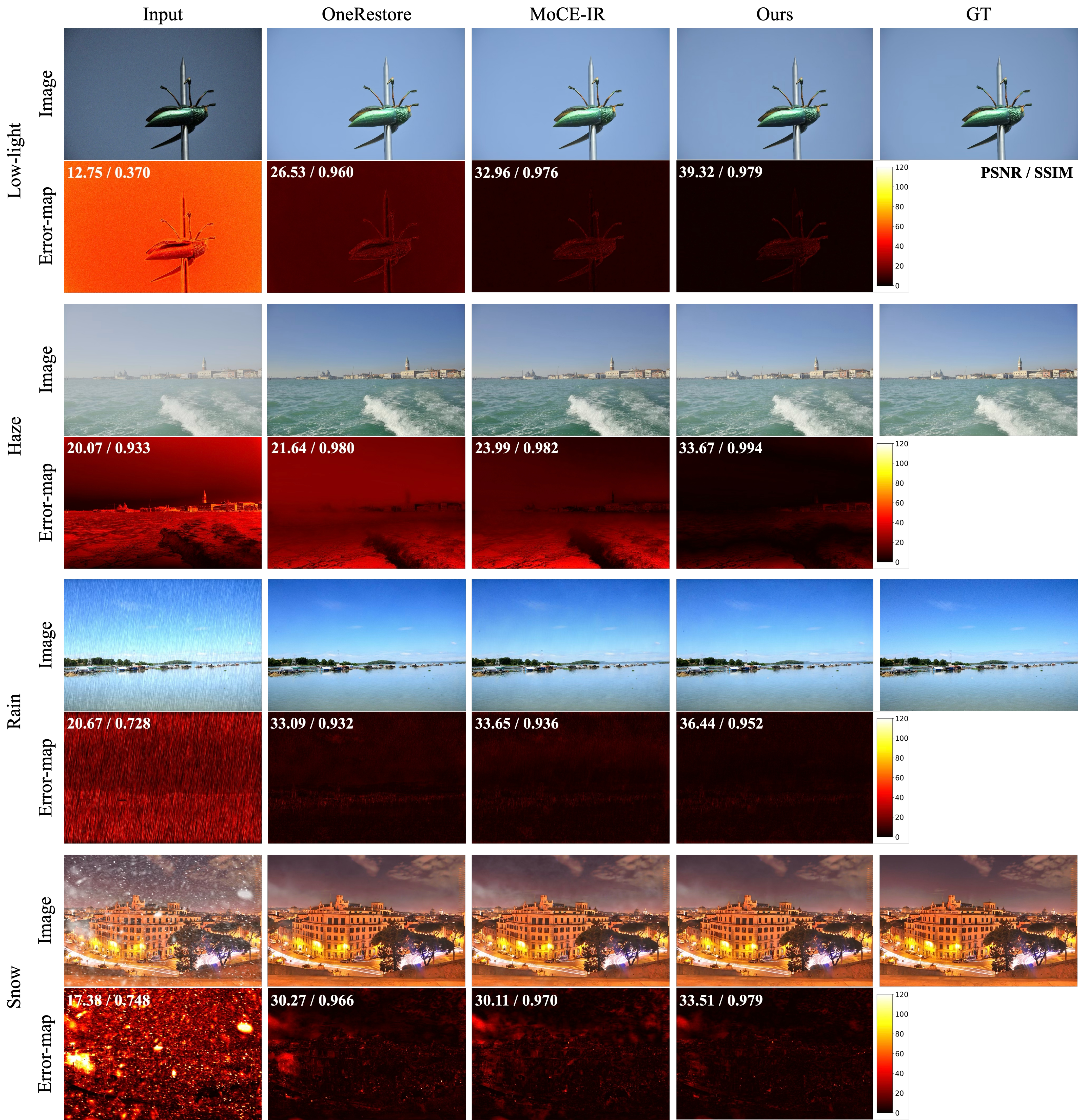}
    \caption{
\textbf{Qualitative comparison on single degradation from the CDD11 dataset.} 
The examples correspond to (from top to bottom): Snow, Rain, Haze, and Low-light. 
For each sample, the top row displays the restored image, while the bottom row visualizes the error maps between the restored output and the ground truth. 
Darker regions in the error maps indicate lower pixel-wise reconstruction errors. 
Our method consistently restores fine details and accurate colors with minimal residual errors compared to OneRestore~\cite{guo2024onerestore} and MOCE-IR~\cite{zamfir2025moceir}.
}
    \label{fig:supp_cdd11_single}
\end{figure}

\cref{fig:supp_cdd11_single} presents visual comparisons under single degradation conditions.
While OneRestore~\cite{guo2024onerestore} and MoCE-IR~\cite{zamfir2025moceir} exhibit color distortions (haze and low-light) and artifacts (rain and snow), our method consistently restores clear details.
This superiority is further visually presented by the error maps.
Our approach yields significantly darker error maps compared to other state-of-the-art methods, indicating much lower pixel-wise reconstruction errors.
Particularly in the haze scenario shown in \cref{fig:supp_cdd11_single}, distinct frequency characteristics coexist: the haze primarily affects the low-frequency global contrast, while the water waves contain high-frequency textures.
While competing methods focus on removing the low-frequency haze component—often discarding fine details like water waves—our method effectively addresses both frequency bands simultaneously.
This capability stems from our MoFE module, which is modulated by the MLLM.
Instead of treating degradations as discrete categories, the MLLM captures their continuous characteristics and intensities.
This fine-grained representation enables our model to precisely target the specific degradation patterns (haze) while preserving the inherent high-frequency details (water waves).

\begin{table}[t]
\centering
\caption{Effect of excluding low-frequency component ($\mathbf{F}^l_{\text{LL}}$) in MoFE.}
\label{tab:abl-low-freq}
\resizebox{0.5\linewidth}{!}{
\begin{tabular}{l|cc|c}
\toprule
  & PSNR & SSIM & \#~Param (M) \\
\midrule
Without $\mathbf{F}^l_{\text{LL}}$ (Ours) & 30.40 & 0.889 & 37.5\\
With $\mathbf{F}^l_{\text{LL}}$   & 30.38 & 0.890 & 38.1 \\
\bottomrule
\end{tabular}
}
\end{table}

\subsection{Additional Qualitative Results on Double Degradation}
\begin{figure}[t]
    \centering
    \includegraphics[width=\linewidth]{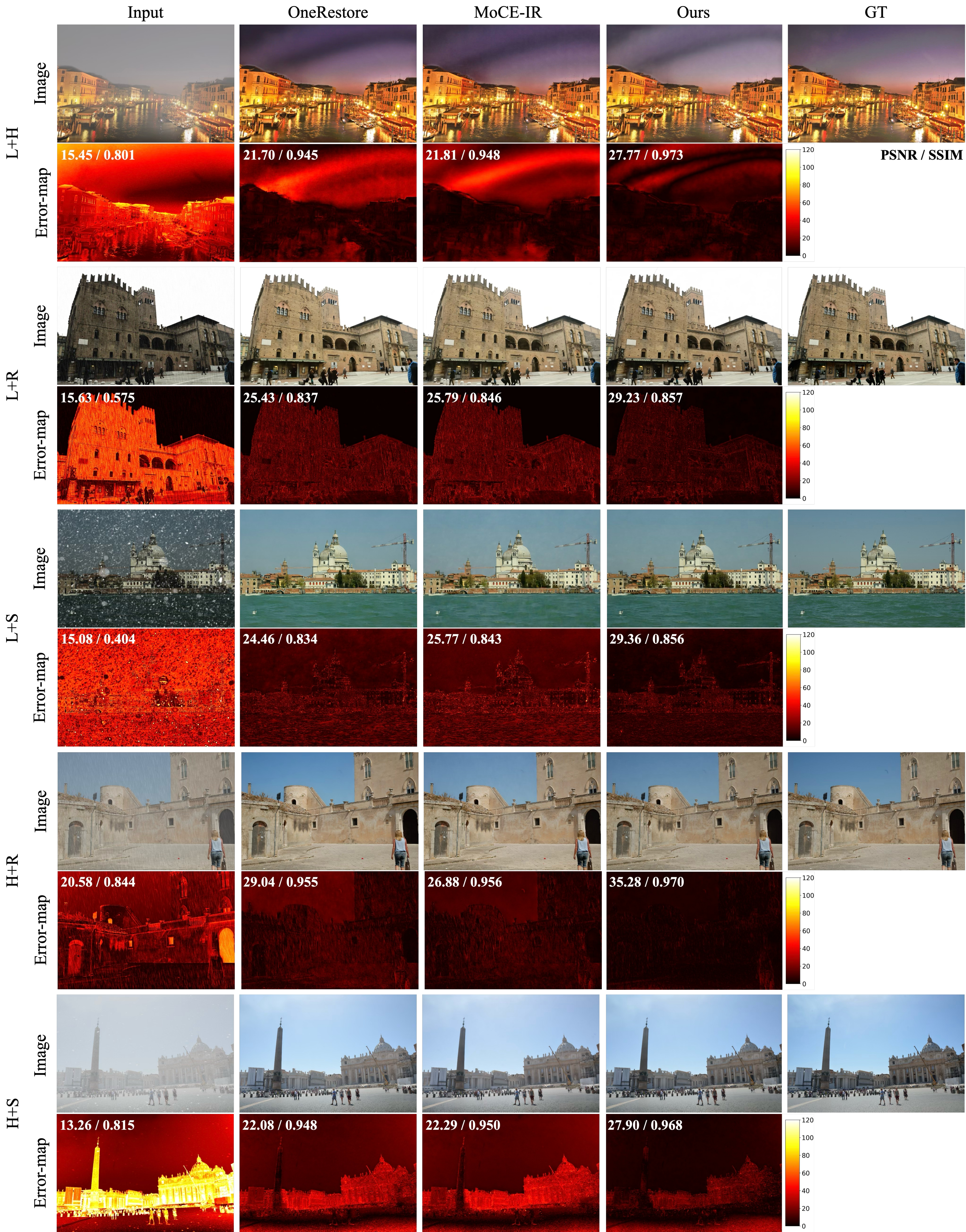}
    \caption{
\textbf{Qualitative comparison on double degradation combinations.} 
The rows represent different composite degradation scenarios.
Our method robustly handles diverse degradation combinations and frequency bands.
}
    \label{fig:supp_cdd11_double}
\end{figure}

We extend our evaluation to double degradation scenarios, where distinct degradation types often coexist, influencing each other during the restoration process.
As shown in \cref{fig:supp_cdd11_double}, OneRestore and MoCE-IR fail to fully address such composite degradations, exhibiting noticeable residual errors distributed across various frequency bands.
In contrast, our method demonstrates a superior capability to handle degradations across a wide frequency spectrum.
This advantage is particularly evident in the haze+rain (H+R) scenario, which presents a challenging mixture of low-frequency haze and high-frequency rain streaks.
While competing approaches tend to leave artifacts in both frequency domains (\eg, sky and building), our method effectively eliminates these components to recover the scene structure with high fidelity.
We attribute this capability to the MoFE module, which leverages diverse frequency experts under the guidance of the MLLM to effectively address scenarios where distinct frequency components coexist.

\subsection{Effect of Including the Low-Frequency Component in MoFE}
Following the design choice in Sec.~3.2, we conduct an ablation study to validate excluding the low-frequency component $\mathbf{F}^l_{\text{LL}}$ from MoFE.
\cref{tab:abl-low-freq} compares models trained with and without explicitly processing $\mathbf{F}^l_{\text{LL}}$.
Including $\mathbf{F}^l_{\text{LL}}$ results in negligible performance differences (PSNR: 30.40 vs.\ 30.38; SSIM: 0.889 vs.\ 0.890), while increasing the number of parameters from 37.5M to 38.1M. 
This suggests that explicitly modeling the low-frequency component is parameter-inefficient and provides no meaningful benefit, which is consistent with our design choice.

\end{document}